%% file: root.tex
\documentclass[letterpaper, 10 pt, conference]{ieeeconf}

\usepackage[utf8]{inputenc}
\usepackage[T1]{fontenc}
\usepackage{mathptmx}
\usepackage{amsmath,amssymb,mathtools,bm}
\usepackage{graphicx}
\usepackage{subcaption}
\usepackage{booktabs}
\usepackage{multirow}
\usepackage[table]{xcolor}
\definecolor{mygray}{RGB}{220,220,220}
\usepackage{siunitx}
\usepackage[linesnumbered,ruled,vlined]{algorithm2e}
\usepackage{url}
\newcommand{\best}[1]{\ifmmode\bm{#1}\else\textbf{#1}\fi}

\newcommand{\unk}{\mathsf{unk}}

\newcommand{\Ind}[1]{\mathbf{1}\{#1\}}
\newcommand{\up}{\textcolor{green!60!black}{$\uparrow$}}
\newcommand{\down}{\textcolor{red!80!black}{$\downarrow$}}
\usepackage{amsfonts} %

\definecolor{darkgray}{gray}{0.35}
\definecolor{lightblue}{HTML}{F3F7FC}
\definecolor{lightmauve}{RGB}{230,220,245}
\IEEEoverridecommandlockouts                              

\title{\LARGE \bf
VORL-EXPLORE: A Hybrid Learning Planning Approach to Multi-Robot Exploration in Dynamic Environments
}

\author{
Ning Liu$^{1}$, Sen Shen$^{2}$, Zheng Li$^{1}$, Sheng Liu$^{3}$,
Dongkun Han$^{2}$, Shangke Lyu$^{4,*}$, and Thomas Br{\"a}unl$^{1}$%
\thanks{This work was supported by the National Natural Science Foundation of China under Grant 62573362.}%
\thanks{$^{*}$Corresponding author. Email: {\tt\small shangke\_lyu@nju.edu.cn}.}%
\thanks{$^{1}$Department of Electrical, Electronic and Computer Engineering, The University of Western Australia, Australia.}%
\thanks{$^{2}$Department of Mechanical and Automation Engineering, The Chinese University of Hong Kong, Hong Kong SAR, China.}%
\thanks{$^{3}$Institute of Material Handling and Logistics, Karlsruhe Institute of Technology, Germany.}%
\thanks{$^{4}$School of Robotics and Automation, Nanjing University, China.}%
}

\begin{document}

\AddToHookNext{shipout/foreground}{%
  \begingroup
  \setlength{\unitlength}{1pt}%
  \put(0,-28){\makebox[\paperwidth][c]{\normalfont\small Accepted at IROS 2026}}%
  \endgroup
}

\maketitle
\thispagestyle{empty}
\pagestyle{empty}

\begin{abstract}
Hierarchical multi-robot exploration decouples frontier allocation from local navigation, making systems brittle in dynamic environments. Without awareness of execution difficulty, allocators may send robots toward frontiers whose routes are congested or unsafe, causing bottlenecks, oscillatory replanning, and redundant coverage. We propose VORL-EXPLORE, a hybrid learning and planning framework that makes exploration execution-aware through execution fidelity, an estimate of local navigability and execution reliability. This fidelity score is used to penalize risky or crowded frontiers in a fidelity-coupled Voronoi allocation objective with inter-robot repulsion, and to arbitrate between global $A^*$ guidance and a reactive reinforcement learning policy during navigation. As a result, robots follow efficient global plans in reliable regions and rely more on safer local reactions in crowded or changing areas. The fidelity model is recalibrated using self-supervised pseudo-labels derived from recent progress and safety outcomes. Quantitative experiments on randomized dynamic grids and an ablation under severe traffic show higher success rates, shorter exploration lengths, and lower inter-robot coverage overlap. A Gazebo factory scenario is included as a deployment illustration. The source code is available at \url{https://github.com/21ning/VORL-EXPLORE}.

\end{abstract}

\section{Introduction}
Large-scale multi-robot systems enable exploration and mapping in dynamic environments with incomplete or changing prior information, such as disaster-response scenes, construction sites, and industrial logistics spaces with moving obstacles, temporary blockages, and time-varying traversability~\cite{zhu2024survey,liu2025cooperative}. Routes that are feasible at one moment can become temporary bottlenecks later. Many exploration frameworks therefore adopt a hierarchical structure that separates global task allocation from local motion execution~\cite{zhou2021fuel}. Under an ideal synchronized shared-state assumption, this design scales well because robots compute assignments and motion actions distributively.
\begin{figure}[!htbp]
    \centering
    \includegraphics[height=2in]{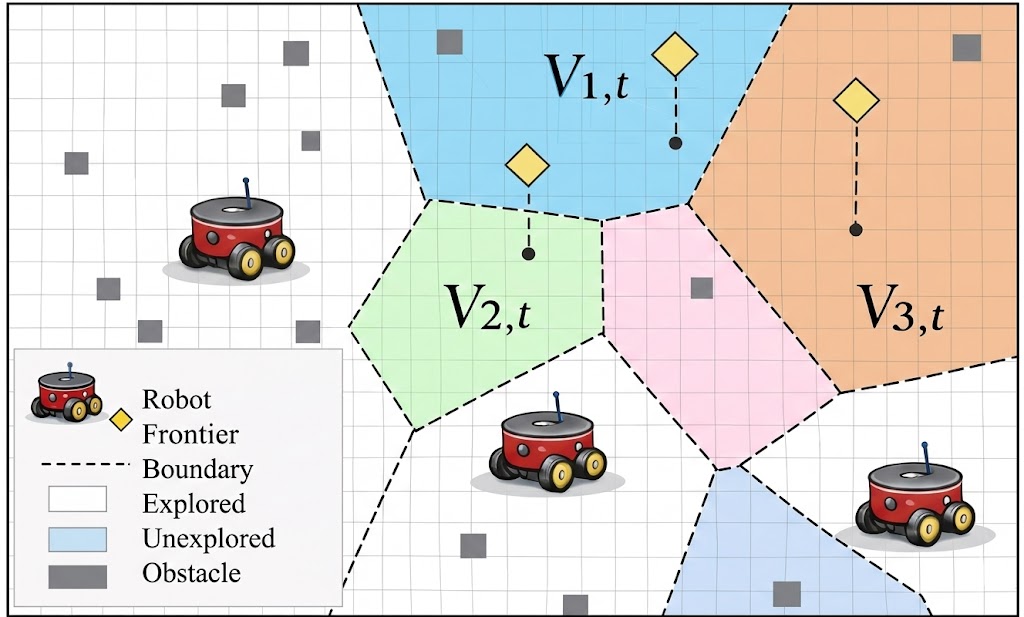}
    \caption{Canonical Voronoi-based frontier assignment in multi-robot exploration. Three robots share an occupancy map, extract frontiers, and form a Voronoi partition $\{V_{i,t}\}$ induced by BFS distance. Each robot selects a frontier within its region as the next exploration goal.}
    \label{fig:Vonori}
\end{figure}
However, the hierarchy becomes fragile when execution difficulty changes faster than the allocator can react. Fig.~\ref{fig:Vonori} provides a minimal static example, where robots select frontiers using distance-driven region partitioning, such as Voronoi rules induced by Breadth-First Search (BFS) distance. Distance-driven allocators work well in open space, but they can dispatch multiple robots toward adjacent frontiers that rely on the same narrow passages, which creates congestion during execution~\cite{pierson2020wbvc}. Robots may then enter yielding oscillations and mutual blocking, repeatedly triggering local replanning~\cite{tordesillas2022mader}. In dynamic environments, this decoupling is amplified because moving obstacles and shifting traffic can invalidate previously good routes, while the upstream objective remains unchanged and keeps regenerating contested patterns. What is missing is a shared execution-fidelity signal that can be updated online and fed back to the allocator, so that target utilities reflect instantaneous navigability and interaction risk~\cite{liang2024hdplanner}.

This hierarchy implicitly assumes that each robot can execute its assigned target reliably under local dynamics~\cite{niu2021vision,huang2024multi}. 
As team size grows and obstacle motion increases, congestion and multi-robot interactions can slow or even stall local navigation. 
These execution failures do not remain local. They cascade through shared corridors, induce mutual blocking, and trigger repeated replanning~\cite{cai2024transformer}. 
Meanwhile, frontier allocators often operate without an online difficulty signal, and local controllers seldom report execution confidence upstream. 
This mismatch can dispatch robots to inefficient targets. It can also amplify path conflicts and redundant coverage~\cite{lajoie2020door}.

Existing attempts to improve robustness often focus on the execution layer, using decentralized learning for collision avoidance or heuristic arbitration for short-horizon decisions~\cite{damani2021primal}. Without bottom-up feedback, the allocator cannot react to local congestion, which leads to redundant travel and deadlocks in high-density settings~\cite{chen2024soscheduler}.

To address this limitation, we propose \textsc{VORL-Explore}, a hybrid learning-and-planning framework for execution-aware multi-robot exploration in dynamic environments. The target problem is not only to assign informative frontiers efficiently, but also to account for whether the assigned routes can be reliably executed under current congestion and obstacle motion. To this end, each robot estimates an execution-fidelity score from local occupancy, teammate congestion, and recent progress cues. The score is fed back to the allocation layer to penalize frontiers whose routes are likely to traverse crowded or unreliable regions. The same score also arbitrates between global A$^\ast$ guidance and a reactive learned policy, allowing robots to follow efficient global plans in reliable regions and rely more on local reactions in congested areas. Finally, the fidelity estimator is recalibrated online using self-supervised pseudo-labels derived from recent coverage gains and safety outcomes.

The fundamental contributions are summarized as follows:
\begin{itemize}
\item A bidirectional closed-loop architecture is proposed to address the structural limitations of strictly top-down multi-robot exploration, unifying the task and motion layers through real-time bottom-up feedback.
\item Execution fidelity is formulated as a shared, continuous representation of local navigability. This single shared signal acts as the architectural link that simultaneously modulates the macroscopic Voronoi-based task assignment and governs the microscopic motion strategy arbitration.
\item A self-supervised online adaptation scheme uses observed progress and safety outcomes to update the fidelity estimator in real time. Together with a deterministic symmetry-breaking recovery rule, it improves execution robustness without requiring manual labels or stationary-environment assumptions.
\end{itemize}

\section{Related Work}
\subsection{Target Assignment and Regional Partitioning}
To scale frontier-based exploration \cite{yamauchi1998frontier} to large teams, decentralized assignment commonly relies on geometric partitioning, such as Voronoi cells \cite{bi2023cure}, or market-based bidding \cite{zlot2002multi}. Most allocators optimize map-based utility using simplified travel cost surrogates \cite{liu2013optimal}. In dense teams and dynamic environments, effective execution cost can vary rapidly due to congestion, multi-robot interaction, and intermittent blockage. Several distributed exploration systems acknowledge these effects but do not explicitly provide a real-time feasibility or congestion signal to the assignment objective \cite{masaba2021gvgexp}. As a result, assignments may remain misaligned with the instantaneous navigability of the environment, which can increase redundant travel and coordination conflicts in narrow passages and crowded regions.

\subsection{Path Planning and Motion Execution}
While classical search-based planners like A* and D* variants provide completeness under static assumptions \cite{koenig2002d, bagad2024optimizing}, they incur substantial replanning overhead and latency in dynamic settings with frequent obstacle motion or multi-robot interactions \cite{ferner2013odrm}.

Learning-based reactive planners improve local efficiency by mapping observations to collision-avoidance actions, and have been widely studied for multi-robot navigation and MAPF-like settings \cite{fan2020distributed}. These methods are effective for short-horizon safety and progress, but they typically operate as local policies that do not expose an explicit execution confidence estimate to the task assignment layer. In addition, policies trained offline can degrade under deployment shifts such as unseen obstacle behaviors or density regimes \cite{ma2021distributed, chen2019end}.
This separation makes it difficult for the system to adjust macroscopic target priorities based on instantaneous local controllability.

\subsection{Connecting Assignment and Execution}
Many systems preserve a cascaded hierarchy where allocation and execution are designed separately \cite{cao2021multi}, which supports scalability but may lead to deadlocks or stalled coverage in narrow corridors and dense interactions \cite{ye2023toward}.
Several works improve robustness within the execution layer via hybrid navigation and arbitration, for example by switching between a classical planner and a learned policy using rule-based detectors \cite{sharma2024hybrid}.
Other approaches learn coordination or communication structures for multi-robot navigation and pathfinding, often leveraging priorities from classical planners to organize local interactions \cite{li2022multi}.
However, the signals used for arbitration or coordination are rarely formulated as a shared and continuously updated variable that can directly modulate frontier assignment.
Our method differs by using an online estimate of local navigability as a coupling signal that influences both frontier scoring and motion arbitration.

\subsection{Online Adaptation in Exploration}
Online adaptation is important when obstacle motion and interaction density shift over time.
Prior work studies self-supervised updates for local collision avoidance and behavior adaptation \cite{kahn2021badgr}, as well as hierarchical navigation that combines global guidance with reactive local controllers \cite{wang2020mobile, faust2018prm}.
Most existing adaptation refines local execution behavior, while the coupling used by the allocator is fixed.
We instead calibrate the coupling signal online using progress and safety feedback to maintain consistency under nonstationarity.
\section{Problem Definition}
\subsection{Environment Model}
A team of $N$ robots explores an initially unknown grid workspace.
Time is discrete with $t\in\mathbb{N}$ and $t=0$ at initialization.
Let $V$ be the set of cells and $E$ be the four-neighbor adjacency edges.
Robot $i$ is at $x_{i,t}\in V$ and selects an action $a_{i,t}\in\mathcal{S}_{i,t}$, where $\mathcal{S}_{i,t}$ contains collision-free moves that satisfy map constraints.
The environment is represented by an occupancy map $M_t:V\rightarrow\{\mathrm{free},\mathrm{occ},\mathrm{unk}\}$. Each robot senses within radius $R_s$ and exchanges pose and map updates to maintain a synchronized occupancy map. This shared-map assumption is commonly adopted in frontier-based exploration systems and enables distributed decision making without a centralized optimizer. We do not model delayed, lossy, or bandwidth-limited communication in this paper.

Frontiers are free cells adjacent to at least one unknown cell,
$\mathcal{F}_t=\{v\in V\mid M_t[v]=\mathrm{free},\exists u\in\mathcal{N}_4[v],M_t[u]=\mathrm{unk}\}$. Dynamic obstacles induce nonstationarity. Let $\mathcal{O}_t\subseteq V$ denote cells occupied by dynamic obstacles at time $t$, moving with speed ratio $\nu\in(0,1)$ relative to robots. Robots treat $\mathcal{O}_t$ as time-varying hard constraints within sensing range. The fused map is updated by overwriting newly observed cells and marking $\mathcal{O}_t$ as $\mathrm{occ}$.

\subsection{Objective}
Let $t^\star$ be the first time when no frontiers remain,
$t^\star=\inf\{t\in\mathbb{N}\mid \mathcal{F}_t=\varnothing\}$.
Define the team-level coverage multiplicity
$\kappa[v]=\left|\left\{i\mid \exists\,t<t^\star,\ x_{i,t}=v\right\}\right|$.
Redundant overlap is the fraction of visited cells covered by at least two robots,
\begin{equation}
\Omega=\frac{\left|\left\{v\in V\mid \kappa[v]\ge 2\right\}\right|}
{\left|\left\{v\in V\mid \kappa[v]\ge 1\right\}\right|}.
\label{eq:overlap}
\end{equation}
We minimize a weighted sum of completion time and redundancy,
\begin{equation}
J=\alpha t^\star+\lambda_{\Omega}\Omega,\quad \alpha\gg\lambda_{\Omega}\ge 0,
\end{equation}
by designing decentralized policies $\{\pi_i\}_{i=1}^N$ that use the shared global map to compute target selection and motion decisions independently.

\section{Methodology}
\begin{figure*}[t]
    \centering
    \includegraphics[width=\linewidth]{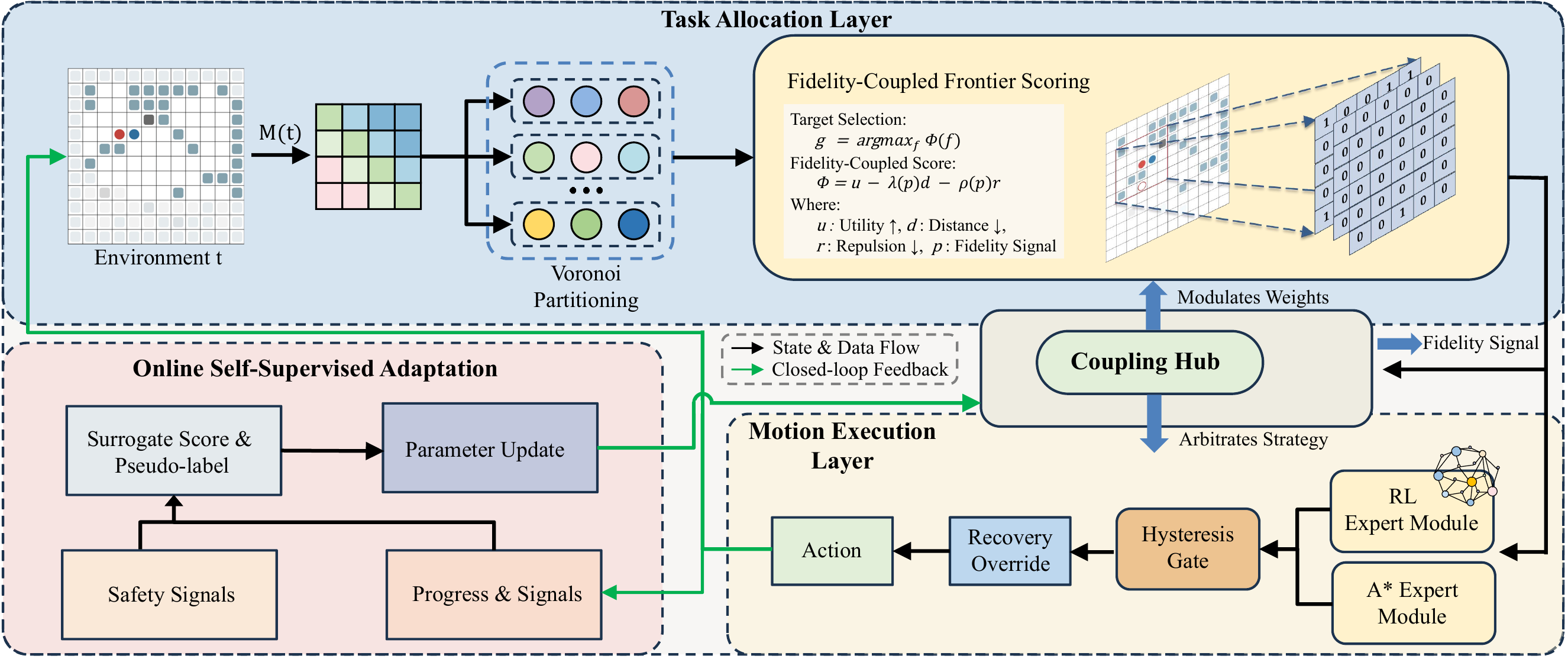}
    \caption{Closed-loop architecture of VORL-EXPLORE. The task-allocation layer extracts frontiers from the shared occupancy map, applies Voronoi filtering, and selects a target using a fidelity-coupled score. The execution-fidelity signal $p_{i,t}$ provides the two coupling links: it modulates the frontier scoring weights in the task layer and controls the hysteresis gate between A$^*$ guidance and the reactive RL policy in the motion layer. Progress and safety outcomes generate pseudo-labels for online adaptation of the fidelity estimator, closing the feedback loop.}
    \label{fig:framework}
\end{figure*}

This work presents a closed-loop multi-robot exploration framework that couples frontier assignment and motion execution through a shared execution-fidelity signal $p_{i,t}$. This signal estimates the local reliability of globally guided motion under current interactions. High fidelity favors informative frontier selection and $A^*$ guidance, whereas low fidelity penalizes risky or congested frontiers and shifts control toward the reactive policy. Progress and safety outcomes update the estimator online, closing the feedback loop between allocation and control.
Fig.~\ref{fig:framework} provides an overview of the architecture and highlights the two coupling links between the task layer and the motion layer.
Let $g_{i,t} \in \mathcal{F}_t \cup \{\emptyset\}$ denote the assigned frontier target for robot $i$ at decision step $t$, and let $p_{i,t}\in[0,1]$ denote the execution-fidelity score used for planner selection.
Two step intervals, $\Delta_{as}$ and $\Delta_{up}$, are used for task reassignment and online parameter updates, respectively.
Task reassignment is executed every $\Delta_{\mathrm{as}}$ decision steps and is additionally triggered immediately when robot $i$ reaches its current goal, i.e., $x_{i,t}=g_{i,t-1}$, or when no valid goal is available ($g_{i,t-1}=\emptyset$). During execution, the reactive policy uses local observations $o_{i,t}$ within a sensing radius $R_s$ (in grid cells), while inter-robot repulsion considers teammates within an interaction radius $R_u$ defined in grid distance.
A fixed window length $W$ maintains the history buffer $\mathcal{H}_{i,t}$ for online adaptation.

\subsection{Closed-loop architecture}
Each robot exchanges map updates with teammates to maintain an approximately consistent shared occupancy map. At each time step, a global frontier candidate set $\mathcal{F}_t$ is extracted. Subsequently, a Voronoi partition induced by the robots' current poses filters this set into a robot-specific subset $\mathcal{F}_{i,t} \subseteq \mathcal{F}_t$. The target $g_{i,t}$ is then assigned by maximizing a fidelity-coupled objective $\Phi_{i,t}[f]$ over $f \in \mathcal{F}_{i,t}$. To evaluate this objective, the shortest-path distance $d_{i,t}[f]$ for each candidate is computed via Breadth-First Search (BFS) on the free-space grid. Concurrently, the robot estimates an execution fidelity score $p_{i,t}$ derived from lightweight local features, including occupancy, congestion, and recent progress. This target assignment is refreshed periodically or triggered asynchronously by significant map updates and recovery activations. Notably, inter-robot repulsion is exclusively evaluated during these reassignment rounds to formulate $\Phi_{i,t}[f]$, rather than during per-step motion execution. The assigned target is then pursued via a fidelity-gated arbitration strategy. The resulting progress and safety outcomes provide continuous self-supervision to update the fidelity estimator online. Algorithm \ref{alg:coex} summarizes this closed-loop procedure, with detailed formulations provided in subsequent sections.

\begin{algorithm}[!htbp]
\small
\DontPrintSemicolon
\caption{Fidelity-Gated Coupled Exploration}
\label{alg:coex}

\KwIn{$x_{i,t}, M_t, o_{i,t}, g_{i,t-1}, \mathcal{H}_{i,t}$; shared $\{x_{j,t}, g_{j,t-1}\}$; $R_s, R_u, \Delta_{\mathrm{as}}, \Delta_{\mathrm{up}}$}
\KwOut{$g_{i,t}$ and $a_{i,t}$}

Update shared state with $(x_{i,t}, g_{i,t-1})$\;
$z_{i,t} \leftarrow \phi(M_t, x_{i,t}, o_{i,t}, \{x_{j,t}\}, \{g_{j,t-1}\}, \mathcal{H}_{i,t})$\;
$p_{i,t} \leftarrow \sigma(w_i^\top z_{i,t} + b_i)$\;
Update $(c^{\mathrm{H}}_{i,t}, c^{\mathrm{L}}_{i,t}, s_{i,t})$\;

\If{$t \bmod \Delta_{\mathrm{as}} = 0$ \textbf{or} $g_{i,t-1}=\emptyset$ \textbf{or} $x_{i,t}=g_{i,t-1}$}{
    $\mathcal{F}_t \leftarrow \textsc{FRONTIERS}(M_t)$; \\
    $g_{i,t} \leftarrow \arg\max_{f\in\mathcal{F}_t} \Phi_{i,t}[f]$; \\
}
\Else{
    $g_{i,t} \leftarrow g_{i,t-1}$\;
}

$(a^{\mathrm{A}}_{i,t}, \textit{ok}) \leftarrow \textsc{PlanA}^\ast(M_t, x_{i,t}, g_{i,t})$\;

\If{$\neg \textit{ok}$}{
    $a^{\mathrm{A}}_{i,t} \leftarrow$ safe no-op\;
    $s_{i,t} \leftarrow 0$\;
}

$a^{\mathrm{E}}_{i,t} \leftarrow \textsc{RL}(o_{i,t})$\;

\If{$s_{i,t}=1$}{
    $\tilde{a}_{i,t} \leftarrow a^{\mathrm{A}}_{i,t}$\;
}
\Else{
    $\tilde{a}_{i,t} \leftarrow a^{\mathrm{E}}_{i,t}$\;
}

$a_{i,t} \leftarrow \textsc{RecoveryOverride}(\tilde{a}_{i,t}, \mathcal{H}_{i,t})$\;

\If{$t \bmod \Delta_{\mathrm{up}} = 0$}{
    Compute $(Q_{i,t}, \tilde{y}_{i,t})$\;
    Update $(w_i, b_i)$\;
}

\end{algorithm}
\subsection{Coupled frontier assignment}
The task layer selects a frontier target from $\mathcal{F}_t$ while accounting for execution reliability.
Each frontier $f$ is associated with an exploration utility $u_t[f]$ that reflects expected information gain.
$u_t[f]$ is defined as the number of unknown cells within a sensing-radius neighborhood around $f$,
\begin{equation}
u_t[f]=\left|\left\{v\in \mathcal{B}(f,R_s)\mid M_t[v]=\unk\right\}\right|,
\end{equation}
where $\mathcal{B}(f,R_s)$ denotes grid cells within radius $R_s$ of $f$.

The assignment also accounts for travel cost and interaction risk. We introduce a repulsion penalty $r_{i,t}[f]$ to discourage assigning multiple robots to nearby frontiers that may trigger local crowding and corridor conflicts. Execution fidelity $p_{i,t}$ continuously modulates the balance among utility, distance, and repulsion. When $p_{i,t}$ is high, $\lambda(\cdot)$ and $\rho(\cdot)$ remain small and the utility term dominates the trade off. When $p_{i,t}$ is low, the distance and repulsion penalties increase, which biases the assignment toward nearby and low conflict frontiers under congestion and dynamic obstacles. The coupled score is

\begin{equation}
\Phi_{i,t}[f]=u_t[f]-\lambda(p_{i,t})\,d_{i,t}[f]-\rho(p_{i,t})\,r_{i,t}[f].
\label{eq:assign_score}
\end{equation}
Within each reassignment round, we apply min--max normalization to $u_t[f]$, $d_{i,t}[f]$, and $r_{i,t}[f]$ over the current frontier set before computing $\Phi_{i,t}[f]$.

To reduce corridor conflicts, the repulsion penalty of assigning robot $i$ to frontier $f$ is defined as
\begin{equation}
\begin{split}
r_{i,t}[f]
&=\frac{1}{N-1}\sum_{j\ne i}\Ind{d_t(f,x_{j,t})\le R_u}
\exp\!\Bigl(-\frac{d_t(f,x_{j,t})}{\sigma_x}\Bigr)\\
&\quad+\frac{\beta}{N-1}\sum_{j\ne i}\Ind{d_t(f,g_{j,t-1})\le R_u}
\exp\!\Bigl(-\frac{d_t(f,g_{j,t-1})}{\sigma_g}\Bigr),
\end{split}
\label{eq:repulsion}
\end{equation}
where $d_t(\cdot,\cdot)$ denotes the same BFS shortest-path distance used to compute $d_{i,t}[f]$ on the current free-space grid induced by $M_t$.
For the target-based term, the indicator is treated as false when $g_{j,t-1}=\emptyset$, so that contribution is zero.
$\sigma_x$ and $\sigma_g$ control the interaction range and $\beta$ balances pose-based and target-based repulsion.

The weights increase as fidelity decreases and a simple linear form is used
\begin{equation}
\lambda(p_{i,t})=\lambda_0+\lambda_1(1-p_{i,t}),\quad
\rho(p_{i,t})=\rho_0+\rho_1(1-p_{i,t}).
\end{equation}
The robot selects the frontier with maximal $\Phi_{i,t}[f]$ and re-evaluates upon periodic refresh or event triggers. This design removes the fragile assumption of reliable execution and injects local feasibility into allocation through a shared continuous signal.

\subsection{Motion arbitration with a learnable switch}
\label{sec:motion_arbitration}

The execution layer produces two complementary actions. A planner generates a globally guided action $a^{\mathrm{A}}_{i,t}$ toward the assigned target using the shared map, while a reactive policy outputs $a^{\mathrm{E}}_{i,t}$ directly from local observations. A lightweight learnable gate outputs an execution-fidelity score $p_{i,t}$, indicating whether globally guided execution is expected to remain reliable under the current local dynamics. The gate is implemented as logistic regression to support decentralized online adaptation,
\begin{equation}
p_{i,t}=\sigma\!\left(w_i^\top z_{i,t}+b_i\right), \qquad z_{i,t}\in\mathbb{R}^8,
\label{eq:gate}
\end{equation}
where $\sigma(\cdot)$ denotes the sigmoid function. The feature vector is produced by a deterministic extractor
\begin{equation}
z_{i,t}=\phi\!\left(M_t, x_{i,t}, o_{i,t}, \{x_{j,t}\}_{j\ne i}, \{g_{j,t-1}\}_{j\ne i}, \mathcal{H}_{i,t}\right)\in\mathbb{R}^{8},
\end{equation}
where $\phi(\cdot)$ returns a compact set of interpretable statistics computed from the occupancy and unknown grids, the feasibility filter, shared teammate states, and a short history buffer $\mathcal{H}_{i,t}$ over the most recent $W$ steps. Concretely, $z_{i,t}$ consists of exactly eight lightweight statistics: local crowding measured by the count of nearby teammates within radius $R_u$ in the shared state, a short-horizon stuck flag, normalized distance to the assigned frontier, feasible-action ratio, unknown-area ratios around the robot and around the goal, local blockage density from four-neighbor free cells, and a planner-feasibility flag indicating whether the $A^*$ branch returns a valid action.

A binary switch state $s_{i,t}\in\{0,1\}$ is used, where $s_{i,t}=1$ selects the planner branch and $s_{i,t}=0$ selects the reactive policy. To reduce oscillatory switching, the hysteresis mechanism uses two thresholds, $\tau_{\mathrm{H}}$ and $\tau_{\mathrm{L}}$, with $\tau_{\mathrm{H}}>\tau_{\mathrm{L}}$, together with a dwell length $K$.
Two counters accumulate consecutive threshold satisfaction,
\begin{equation}
c^{\mathrm{H}}_{i,t}=\mathbb{I}\{p_{i,t}\ge\tau_{\mathrm{H}}\}\bigl(c^{\mathrm{H}}_{i,t-1}+1\bigr),\quad
c^{\mathrm{L}}_{i,t}=\mathbb{I}\{p_{i,t}\le\tau_{\mathrm{L}}\}\bigl(c^{\mathrm{L}}_{i,t-1}+1\bigr),
\label{eq:hyst_counter}
\end{equation}
and the switch updates only after $K$ consistent steps,
\begin{equation}
s_{i,t}=
\begin{cases}
1, & s_{i,t-1}=0 \;\land\; c^{\mathrm{H}}_{i,t}\ge K,\\
0, & s_{i,t-1}=1 \;\land\; c^{\mathrm{L}}_{i,t}\ge K,\\
s_{i,t-1}, & \text{otherwise}.
\end{cases}
\label{eq:switch_state}
\end{equation}
This yields a preliminary action
\begin{equation}
\tilde a_{i,t}=
\begin{cases}
a^{\mathrm{A}}_{i,t}, & s_{i,t}=1,\\
a^{\mathrm{E}}_{i,t}, & s_{i,t}=0.
\end{cases}
\label{eq:action_select}
\end{equation}

Finally, a recovery override enforces safety and breaks stagnation. It triggers on planning infeasibility, stalled progress over $W$ steps, or frequent switch oscillation, executes a short symmetry-breaking maneuver, and then resumes normal gated execution.

\subsection{Online self-supervised adaptation}
The gate is updated online without manual labels so that the switch remains calibrated in unseen environments. After execution, a surrogate quality score $Q_{i,t}$ is computed over a short sliding window of length $W$:
\begin{equation}
Q_{i,t}= \omega_{\mathrm{cov}}\Delta \mathrm{cov}_{i,t}+\omega_{\mathrm{dist}}\Delta \mathrm{dist}_{i,t}-\omega_{\mathrm{risk}}\mathrm{risk}_{i,t}-\omega_{\mathrm{stall}}\mathrm{stall}_{i,t}.
\label{eq:surrogate_score}
\end{equation}
where $\omega_* > 0$ are fixed weights. To match the planning layer, $\Delta \mathrm{dist}_{i,t}$ uses the BFS shortest-path distance on the current occupancy grid, so necessary obstacle detours are not penalized. $\Delta \mathrm{cov}_{i,t}$ counts newly observed cells, $\mathrm{risk}_{i,t}$ aggregates collisions and constraint violations, and $\mathrm{stall}_{i,t}$ penalizes lack of progress.

A pseudo label is generated from the score using the indicator function $\mathbb{I}\{\cdot\}$:
\begin{equation}
\tilde y_{i,t}=\mathbb{I}\{Q_{i,t}\ge 0\}.
\label{eq:pseudo_label}
\end{equation}
The local gate is trained with regularized binary cross entropy loss, where $\lambda_{\mathrm{reg}}$ controls the L2 regularization:
\begin{equation}
\mathcal{L}_{i,t}= -\tilde y_{i,t}\log p_{i,t} - \bigl(1-\tilde y_{i,t}\bigr)\log\{1-p_{i,t}\} + \frac{\lambda_{\mathrm{reg}}}{2}\lVert w_i\rVert_2^2.
\end{equation}
Updates are applied only when the score magnitude exceeds a clear margin $m > 0$ to reduce noise driven drift. A single online step with learning rate $\eta$ is:
\begin{align}
w_i &\leftarrow w_i
-\eta\,\mathbb{I}\{|Q_{i,t}|\ge m\}\Bigl(\{p_{i,t}-\tilde y_{i,t}\}z_{i,t}+\lambda_{\mathrm{reg}} w_i\Bigr), 
\label{eq:gate_update_w}\\
b_i &\leftarrow b_i
-\eta\,\mathbb{I}\{|Q_{i,t}|\ge m\}\{p_{i,t}-\tilde y_{i,t}\}.
\label{eq:gate_update_b}
\end{align}
This mechanism enables the learnable switch to track changing local dynamics and maintain coherent allocation and execution behavior.

\input{table}
\section{Experiments}

\subsection{Experimental Setup and Baselines}
\subsubsection{Training and Environment}
We implement the reactive RL execution policy $\pi^{\mathrm{E}}$ using the \textsc{EPOM} design~\cite{skrynnik2023switch} and train it with PPO using a learning rate of $1\times10^{-4}$ and discount factor $\gamma=0.99$.
All policy parameters are trained from scratch on procedurally generated $64\times64$ grids with $N=64$ robots and a horizon of 512 steps.
Static obstacles are sampled independently per cell with occupancy probability $p_{\text{occ}}\sim\mathcal{U}[0.15,0.45]$ to induce large-scale coordination under varying obstacle densities.
We set the sensing radius and interaction radius to $R_s=3$ and $R_u=3$ cells.
Dynamic obstacles move at a fixed speed ratio $\nu=0.5$ relative to the robot speed in all experiments.

The logistic gate is warm-started using self-supervised rollouts collected in the same procedural environments and under the same robot density. In all main comparison experiments, we freeze the gate parameters at test time and use a fixed checkpoint to ensure fair comparison and reproducibility. Test-time gate updates are disabled unless otherwise stated, and are evaluated only in the dedicated severe-traffic ablation of Table~\ref{tab:ablation_online}.

Evaluation uses held-out $40\times40$ and $80\times80$ grids with $30\%$ static obstacle density.
To examine scalability and congestion effects, we vary both the number of dynamic obstacles and the team size across scenarios.
The episode horizon scales with map size to control for size-dependent effects.
All hyperparameters remain fixed during testing.
\subsubsection{Evaluation Protocol and Baselines}
Evaluation uses all methods under a decoupled two-stage exploration pipeline with a unified protocol.
All methods share the same discrete action space, feasibility filter $\mathcal{S}_{i,t}$, horizon $T_{\max}$, and reassignment schedule.
At the protocol level, we assume ideal shared-state communication to maintain a synchronized occupancy map for frontier extraction and goal assignment.
In executor-level evaluation, all learning-based executors, including \textsc{VORL-Explore}, act only on local observations and do not access the global map or any teammate states. The corresponding executor-level comparisons across map sizes and dynamic-obstacle densities are reported in Table~\ref{tab:results}.

We separate the baselines into allocator-level and executor-level groups because most execution methods operate through a frontier-goal interface. 
In the allocator-level comparison, only the module that selects $g_{i,t}\in\mathcal{F}_t$ is replaced, while the execution module remains fixed to that of \textsc{VORL-Explore}. 
We compare greedy frontier assignment~\cite{yamauchi1997frontier}, Hungarian matching~\cite{kuhn1955hungarian}, and auction-based allocation~\cite{zlot2002multi}.

In the executor-level comparison, the allocator and the reassignment schedule are fixed. 
At each reassignment step, all methods receive the same frontier goal $g_{i,t}$, and only the local execution module that maps $(o_{i,t}, g_{i,t})$ to an action is replaced. 
To enforce a unified interface, actions are filtered by the same feasibility set $\mathcal{S}_{i,t}$, and executors do not access the global map or teammate states. 
We benchmark \textsc{MATS-LP}~\cite{skrynnik2024decentralized}, \textsc{DHC}~\cite{ma2021distributed}, \textsc{PICO}~\cite{li2022multi}, and \textsc{ICBS}~\cite{boyarski2015icbs} using their published hyperparameters, adapting only environment-specific interface parameters such as grid size and resolution, the discrete action set, collision-checking rules, and the episode horizon. 
Finally, we include two ablated variants, VORL-$A^\ast$ and VORL-RL, to isolate the contribution of the gating mechanism. Fig.~\ref{fig:gazebo} further examines simulator-level transfer without additional fine-tuning.
\subsection{Results}

\subsubsection{Target Allocation and Scalability} 
\begin{figure}[!htbp]
    \centering
    \begin{subfigure}[b]{0.48\columnwidth}
        \centering
        \includegraphics[width=\linewidth]{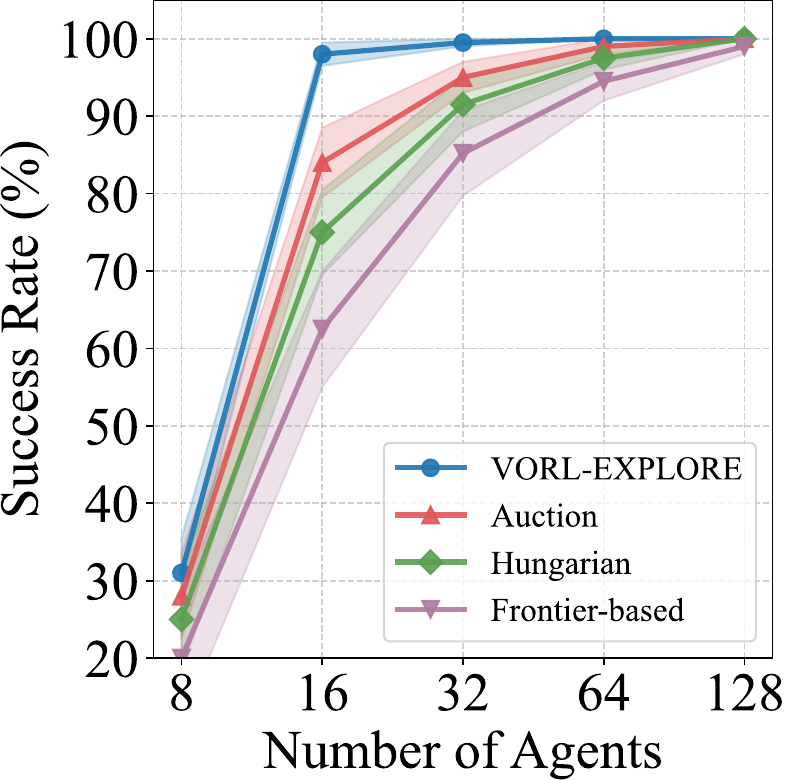}
        \caption{SR}
        \label{fig:success_all}
    \end{subfigure}
    \hfill
    \begin{subfigure}[b]{0.48\columnwidth}
        \centering
        \includegraphics[width=\linewidth]{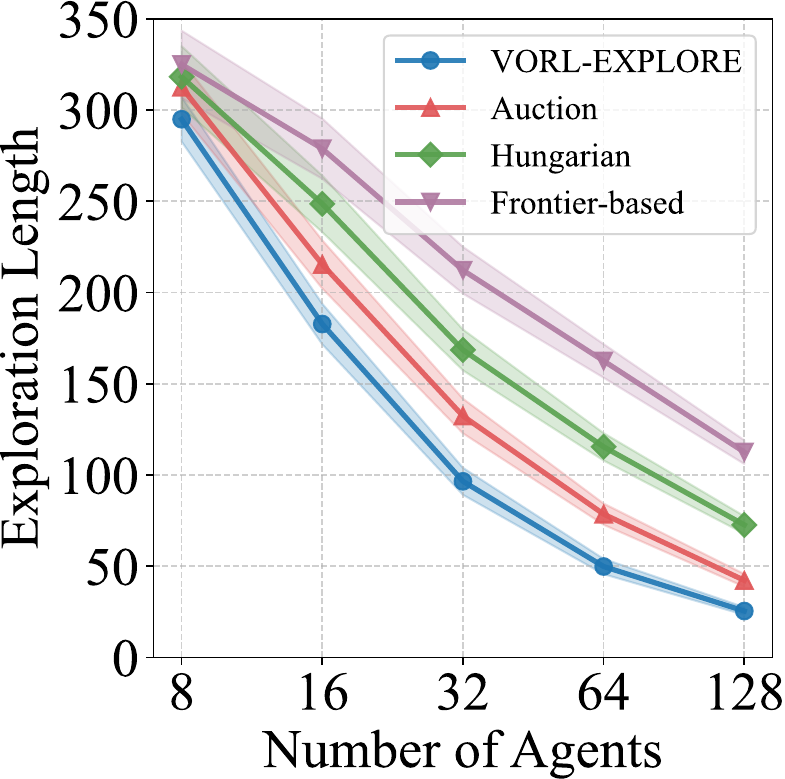}
        \caption{EL}
        \label{fig:ep_len_all}
        \end{subfigure}
        \caption{Allocator-level comparison with target allocation baselines on an $80\times80$ grid with 16 dynamic obstacles and 30\% static obstacle density.
        All methods share the same \textsc{VORL-Explore} execution layer.
        Curves show the mean over 100 runs and shaded regions indicate 95\% confidence intervals.}
    \label{fig:taskallocation}
\end{figure}

We evaluate scalability by varying the team size on an $80 \times 80$ grid with 16 dynamic obstacles and 30\% static obstacle density. Fig.~\ref{fig:taskallocation} reports success rate and exploration length over successful runs. Auction, Hungarian, and frontier based allocators improve as the number of robots increases at first, then show diminishing returns. These methods score targets largely by static travel distance, so they do not account for congestion and moving obstacles during execution in narrow passages. As the team grows, multiple robots are dispatched toward adjacent frontiers, which increases mutual blocking and redundant travel. The frontier based baseline achieves the lowest success rates and shows little reduction in exploration length. Auction is the strongest decoupled baseline but still plateaus at an exploration length of $42.2$ at maximum team size. In contrast, \textsc{VORL-Explore} exhibits rapid and continuous convergence in exploration efficiency. Its EL drops dramatically to $25.4$, and its corresponding SR curve achieves a robust $100\%$ significantly earlier than all tested baselines. By integrating global target assignment with local reinforcement-learning execution, the system dynamically anticipates local congestion. Rather than assigning multiple robots to adjacent frontiers in congested areas, \textsc{VORL-Explore} modulates target selection using execution fidelity, which promotes better spatial dispersion and more consistent marginal gains as team size increases.

\subsubsection{Motion Execution Performance}
Table~\ref{tab:results} compares motion execution on $40\times40$ and $80\times80$ grids under increasing numbers of dynamic obstacles. The table reports values as mean $\pm$ standard deviation. On $40\times40$ with 64 dynamic obstacles, \textsc{VORL-Explore} achieves SR 0.95 with Overlap 0.21, while \textsc{PICO} drops to SR 0.51 with Overlap 0.33. On $80\times80$, the gap widens under dense traffic. \textsc{VORL-Explore} maintains SR 0.96 at 64 dynamic obstacles, whereas \textsc{ICBS} declines from SR 0.85 at 8 obstacles to SR 0.31 at 64 obstacles, and Overlap increases from 0.36 to 0.51. \textsc{PICO} also remains low at SR 0.42 at 64 obstacles, which is consistent with deadlocks in crowded corridors. In contrast, \textsc{VORL-Explore} sustains success rates above 90\% with low redundant revisits across traffic levels, which indicates robust execution when congestion increases and baseline methods degrade. 

Table~\ref{tab:results} also includes two single executor variants to contextualize the hybrid switch. VORL-A$^\ast$ is effective in light traffic but its SR decreases with congestion, reaching 0.55 on the $80\times80$ maps with 64 dynamic obstacles. VORL-RL remains robust at high density and reaches SR 0.92 under the same setting, but it incurs longer exploration in sparse layouts. On $80\times80$ with 8 dynamic obstacles, VORL-RL yields EL 203.20, while \textsc{VORL-Explore} reduces this to EL 181.32. By selecting between planning guidance and reactive avoidance using execution fidelity, \textsc{VORL-Explore} achieves strong performance across both sparse and highly congested regimes. These trends indicate robust execution as traffic increases, without introducing additional online computation beyond the per-step gate evaluation. 

\subsection{Ablation Study}
\label{sec:V.C}

Table~\ref{tab:results} already reports two single-executor variants, VORL-A$^\ast$ and VORL-RL, which enforce planning-only or policy-only execution and therefore isolate the effect of gate-based arbitration. 
Here we focus on the remaining design factors. We progressively enable the two coupling links, CA and CP, in Fig.~\ref{fig:ablation_results}, and we evaluate warm-start initialization and online adaptation of the execution fidelity estimator in Table~\ref{tab:ablation_online} under heavy traffic.

\input{table2}

\subsubsection{Impact of the Coupled Architecture}
We compare four variants, \emph{Base}, Coupled Planning (\emph{CP}), Coupled Assignment (\emph{CA}), and \emph{Full}. \emph{Base} decouples assignment and execution, so the allocator relies on simplified travel costs while the executor must resolve congestion and moving obstacles. \emph{CA} couples assignment with execution fidelity, which downweights frontiers that are likely to induce detours and interference. \emph{CP} enables gate driven switching between planning and the reactive policy, which reduces local deadlocks when crowding patterns change. \emph{Full} enables both links. Fig.~\ref{fig:ablation_results} shows that each link improves success and reduces both exploration length and overlap, and \emph{Full} provides the most stable behavior across runs.

\subsubsection{Effectiveness of Online Adaptation}
We evaluate the gate initialization and online-update strategies in a severe traffic regime on a $40\times40$ grid with 128 dynamic obstacles.
Frequent encounters rapidly change local navigability, making a fixed predictor prone to miscalibration and repeated no-progress behavior.
Table~\ref{tab:ablation_online} reports mean $\pm$ STD over 100 runs. Without pre-training or online updates, the cold static gate attains SR 0.36, triggers recovery 82.4 times per episode, and selects the planner for 91.5\% of steps, indicating over-reliance on planning under congestion.
Warm initialization alone improves robustness, raising SR to 0.51 and reducing recovery to 42.7, but the gate remains fixed and still over-selects planning at 74.2\%. Enabling online updates yields a substantially larger gain. Even without pre-training, the cold adaptive variant increases SR to 0.65 while reducing recovery to 12.3 and planner selection to 37.8\%, showing that test-time updates effectively recalibrate the gate to the current traffic conditions.  Combining warm initialization with online updates further improves SR to 0.69 and achieves the lowest recovery 6.8 with planner selection 34.5\%. Overall, pre-training provides a better starting point, whereas online adaptation is the dominant factor that maintains calibration under non-stationary interactions, reducing both excessive planning and recovery usage.

\begin{table}[!bpht]
    \centering
    \normalsize
    \renewcommand{\arraystretch}{1.4} %
    \caption{Ablation of gate initialization (Pre-train) and test-time gate updates (Online) under severe traffic. Values are mean $\pm$ STD over 100 runs on a $40\times40$ grid with 30\% static obstacles and 128 dynamic obstacles.}
    \label{tab:ablation_online}
    \resizebox{\columnwidth}{!}{
    \begin{tabular}{l cc cccc}
        \toprule
        \multirow{2}{*}{\textbf{Variant}} & \multicolumn{2}{c}{\textbf{Setup}} & \multicolumn{4}{c}{\textbf{Metrics}} \\
        \cmidrule(lr){2-3} \cmidrule(lr){4-7}
        & \textbf{Pre-train} & \textbf{Online} & \textbf{SR}~\up & \textbf{EL}~\down & \textbf{Recov.}~\down & $\mathbf{\pi_{plan}}$ \textbf{(\%)} \\
        \midrule
        
        $\text{VORL}_{\text{Static}}$ (Cold) & $\times$ & $\times$ 
        & 0.36$\pm$0.11 & 248.2$\pm$29.5 & 82.4$\pm$15.2 & 91.5 \\
        
        $\text{VORL}_{\text{Static}}$ (Warm) & \checkmark & $\times$ 
        & 0.51$\pm$0.13 & 224.5$\pm$22.1 & 42.7$\pm$9.6 & 74.2 \\
        
        \midrule
        
        $\text{VORL}_{\text{Adapt}}$ (Cold) & $\times$ & \checkmark 
        & 0.65$\pm$0.07 & 208.7$\pm$17.5 & 12.3$\pm$4.2 & 37.8 \\
        
        \rowcolor[HTML]{f3f7fc}
        \textbf{VORL-EXPLORE} & \textbf{\checkmark} & \textbf{\checkmark} 
        & \textbf{0.69$\pm$0.05} & \textbf{204.3$\pm$14.2} & \textbf{6.8$\pm$2.1} & \textbf{34.5} \\
        
        \bottomrule
    \end{tabular}%
    }
\end{table}
\subsection{Experimental Validation}

\begin{figure}[!bhtp]
    \centering
    \begin{subfigure}[b]{0.48\columnwidth}
        \centering
        \includegraphics[width=\linewidth]{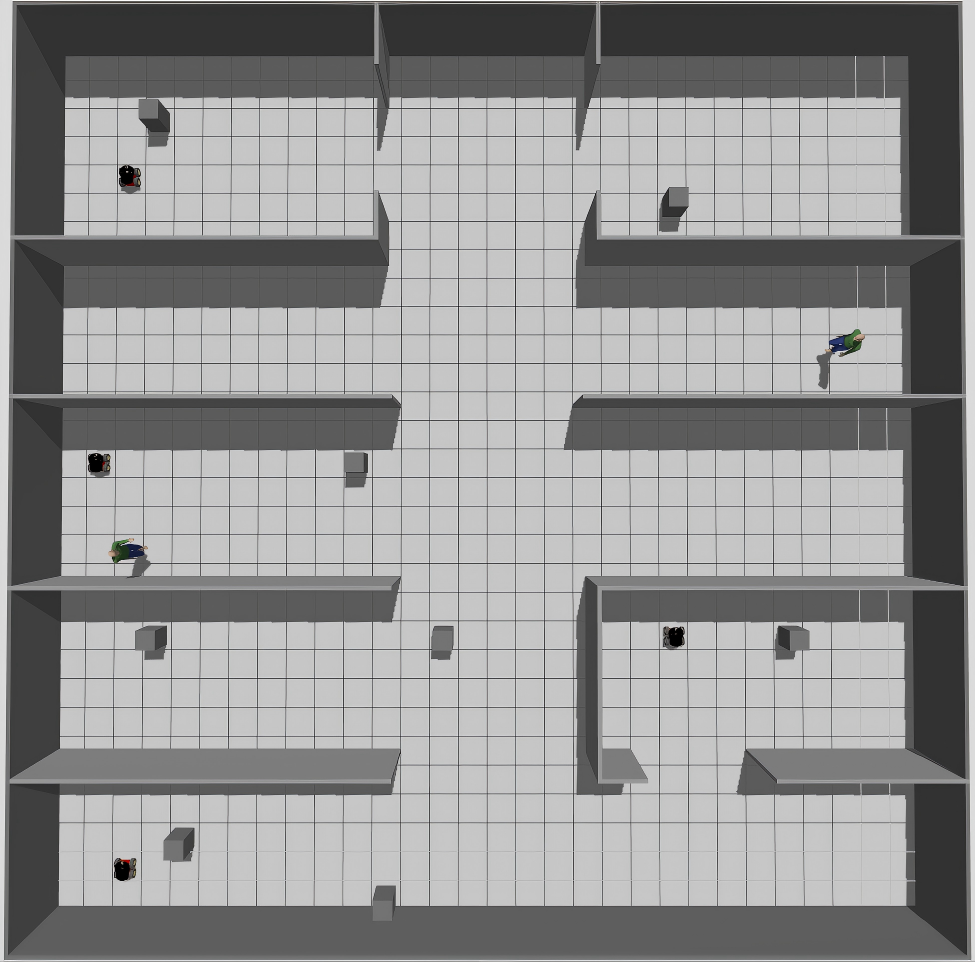}
        \caption{Gazebo factory scene}
        \label{fig:gazebo_sim}
    \end{subfigure}
    \hfill
    \begin{subfigure}[b]{0.48\columnwidth}
        \centering
        \includegraphics[width=\linewidth]{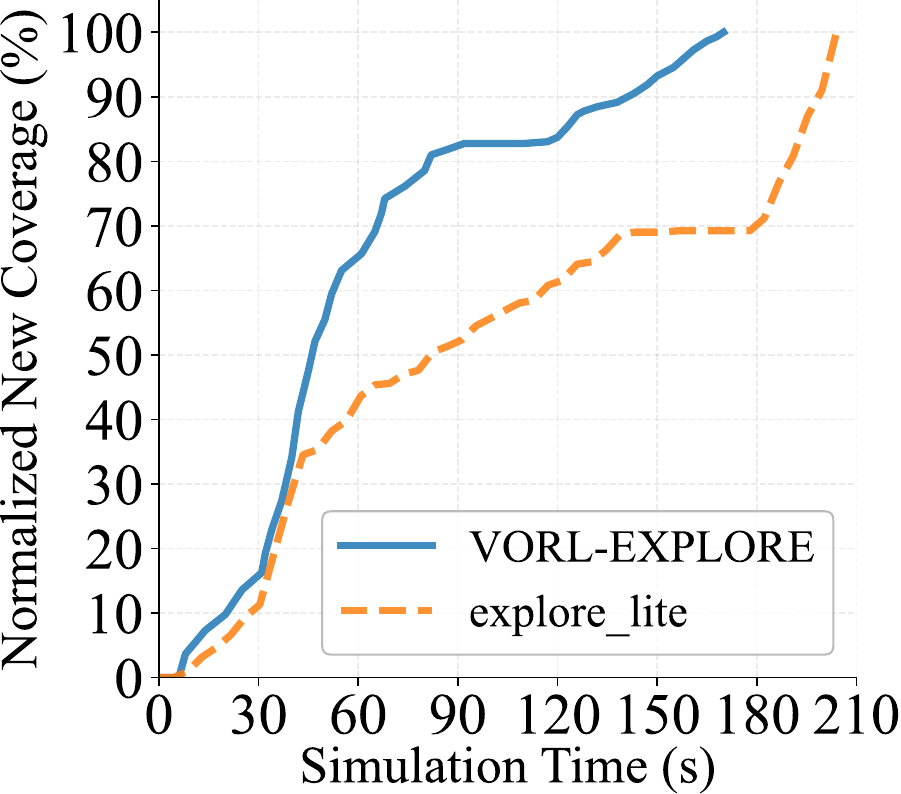}
        \caption{Coverage curve}
        \label{fig:gz_coverage}
    \end{subfigure}
    \caption{Gazebo proof-of-concept study with four Pioneer3 robots in a dynamic factory-like environment. The normalized new coverage curve compares \textsc{VORL-Explore} with ROS explore\_lite under the same simulated conditions.}
    \label{fig:gazebo}
\end{figure}

To demonstrate simulator-level integration, we evaluate four Pioneer3 robots in a factory-like Gazebo environment with static obstacles and two constant-velocity pedestrians. Under the same synchronized shared-state assumption and without additional fine-tuning beyond grid-based training, \textsc{VORL-Explore} continuously expands the explored region. Fig.~\ref{fig:gazebo} shows the Gazebo factory scene and normalized new coverage, defined as the percentage of newly explored cells relative to the final explored area.

Compared with ROS explore\_lite under identical conditions, \textsc{VORL-Explore} achieves earlier coverage gains and maintains a higher coverage rate. This study is intended only as a simulator-level proof of
integration rather than a complete validation of deployment robustness. The Gazebo experiment demonstrates that the proposed coupling mechanism can be integrated into a robotics simulation framework, while more challenging conditions such as communication delay, perception noise, and dense human traffic remain future work.

\section{Conclusion}
This paper presented VORL-EXPLORE, a decentralized hybrid learning-and-planning framework that couples frontier assignment and motion execution through execution fidelity. The fidelity signal modulates frontier scoring and arbitrates between A$^\ast$ guidance and a reactive RL policy. Experiments in dynamic grid worlds show improved success rate, exploration length, and redundancy, while the Gazebo study demonstrates simulator-level integration. The current evaluation assumes synchronized state sharing and simplified sensing. Future work will address communication delay, perception and localization noise, denser traffic, and narrower passages.

\bibliographystyle{./IEEEtran} %
\bibliography{./IEEEexample}

\end{document}

%% file: table.tex
\begin{table*}[t]
    \centering
    \scriptsize
    \caption{Experimental results on $40\times40$ and $80\times80$ grids over 100 runs. Values are reported as mean $\pm$ standard deviation. SR: success rate (\up), EL: Exploration length (\down), and Overlap: redundant overlap ratio (\down). Best values are \textbf{bolded}.}
    \label{tab:results}
    \resizebox{1\textwidth}{!}{%
    \begin{tabular}{l*{12}{c}}
        \toprule
        \multirow{2}{*}{Method}
        & \multicolumn{3}{c}{8 Dynamic obstacles}
        & \multicolumn{3}{c}{16 Dynamic obstacles}
        & \multicolumn{3}{c}{32 Dynamic obstacles}
        & \multicolumn{3}{c}{64 Dynamic obstacles} \\
        \cmidrule(lr){2-4}\cmidrule(lr){5-7}\cmidrule(lr){8-10}\cmidrule(lr){11-13}
        & SR~\up & EL~\down & Overlap~\down
        & SR~\up & EL~\down & Overlap~\down
        & SR~\up & EL~\down & Overlap~\down
        & SR~\up & EL~\down & Overlap~\down \\
        \midrule

        \multicolumn{13}{l}{\textcolor{darkgray}{$\textcolor{blue!42!green}{\bullet}$ 40$\times$40 world, 4 Agents, 30\% static obstacle density}} \\

        \ \  DHC
        & 0.90$\pm$0.08 & 169.22$\pm$11.3 & 0.23$\pm$0.03
        & 0.88$\pm$0.09 & 171.78$\pm$12.1 & 0.23$\pm$0.03
        & 0.87$\pm$0.10 & 179.64$\pm$14.5 & 0.23$\pm$0.03
        & 0.88$\pm$0.09 & 191.45$\pm$15.2 & 0.24$\pm$0.04 \\

        \ \  PICO
        & 0.71$\pm$0.12 & 212.80$\pm$18.4 & 0.32$\pm$0.05
        & 0.62$\pm$0.15 & 213.60$\pm$19.1 & 0.32$\pm$0.05
        & 0.56$\pm$0.14 & 221.40$\pm$20.3 & 0.33$\pm$0.06
        & 0.51$\pm$0.16 & 231.42$\pm$21.5 & 0.33$\pm$0.06 \\

        \ \  MATS-LP
        & 0.85$\pm$0.09 & 198.75$\pm$14.2 & 0.26$\pm$0.04
        & 0.82$\pm$0.10 & 203.12$\pm$15.4 & 0.26$\pm$0.04
        & 0.85$\pm$0.11 & 205.33$\pm$16.1 & 0.27$\pm$0.05
        & 0.87$\pm$0.10 & 207.84$\pm$17.5 & 0.25$\pm$0.04 \\

        \ \ ICBS
        & 0.78$\pm$0.11 & 182.20$\pm$13.1 & 0.27$\pm$0.04
        & 0.68$\pm$0.14 & 194.40$\pm$16.5 & 0.29$\pm$0.05
        & 0.48$\pm$0.16 & 218.30$\pm$22.4 & 0.35$\pm$0.07
        & 0.22$\pm$0.18 & 235.60$\pm$30.2 & 0.41$\pm$0.09 \\

        \ \  VORL-A*
        & 0.91$\pm$0.07 & 170.50$\pm$10.2 & 0.24$\pm$0.03
        & 0.84$\pm$0.09 & 182.30$\pm$14.4 & 0.26$\pm$0.04
        & 0.68$\pm$0.13 & 204.40$\pm$21.5 & 0.31$\pm$0.06
        & 0.44$\pm$0.15 & 225.60$\pm$28.1 & 0.36$\pm$0.08 \\

        \ \  VORL-RL
        & 0.81$\pm$0.10 & 185.20$\pm$15.5 & 0.27$\pm$0.05
        & 0.83$\pm$0.09 & 182.60$\pm$14.2 & 0.26$\pm$0.04
        & 0.87$\pm$0.08 & 178.10$\pm$12.5 & 0.24$\pm$0.04
        & 0.89$\pm$0.08 & 184.50$\pm$13.1 & 0.23$\pm$0.04 \\

        \ \  \cellcolor{lightmauve!40}VORL-EXPLORE
        & \cellcolor[HTML]{f3f7fc}\best{0.94}$\pm$0.06 & \cellcolor[HTML]{f3f7fc}\best{165.35}$\pm$9.8 & \cellcolor[HTML]{f3f7fc}\best{0.21}$\pm$0.02
        & \cellcolor[HTML]{f3f7fc}\best{0.92}$\pm$0.07 & \cellcolor[HTML]{f3f7fc}\best{168.42}$\pm$10.3 & \cellcolor[HTML]{f3f7fc}\best{0.22}$\pm$0.02
        & \cellcolor[HTML]{f3f7fc}\best{0.94}$\pm$0.06 & \cellcolor[HTML]{f3f7fc}\best{174.01}$\pm$10.8 & \cellcolor[HTML]{f3f7fc}\best{0.21}$\pm$0.02
        & \cellcolor[HTML]{f3f7fc}\best{0.95}$\pm$0.05 & \cellcolor[HTML]{f3f7fc}\best{182.39}$\pm$11.9 & \cellcolor[HTML]{f3f7fc}\best{0.21}$\pm$0.02 \\

        \midrule

        \multicolumn{13}{l}{\textcolor{darkgray}{$\textcolor{blue!42!green}{\bullet}$ 80$\times$80 world, 16 Agents, 30\% static obstacle density}} \\

        \ \  DHC
        & 0.99$\pm$0.01 & 186.94$\pm$15.5 & 0.32$\pm$0.04
        & 0.96$\pm$0.03 & 188.37$\pm$16.1 & 0.32$\pm$0.04
        & 0.91$\pm$0.05 & 188.41$\pm$16.2 & 0.33$\pm$0.05
        & 0.87$\pm$0.07 & 193.02$\pm$17.5 & 0.34$\pm$0.05 \\

        \ \  PICO
        & 0.52$\pm$0.14 & 287.24$\pm$25.4 & 0.47$\pm$0.07
        & 0.51$\pm$0.15 & 292.42$\pm$26.8 & 0.48$\pm$0.08
        & 0.49$\pm$0.14 & 293.39$\pm$27.1 & 0.48$\pm$0.07
        & 0.42$\pm$0.16 & 303.52$\pm$30.2 & 0.49$\pm$0.09 \\

        \ \  MATS-LP
        & 0.93$\pm$0.06 & 219.34$\pm$18.5 & 0.36$\pm$0.05
        & 0.91$\pm$0.07 & 222.88$\pm$19.3 & 0.37$\pm$0.06
        & 0.86$\pm$0.08 & 227.27$\pm$20.8 & 0.36$\pm$0.05
        & 0.88$\pm$0.08 & 234.15$\pm$21.4 & 0.36$\pm$0.05 \\

        \ \  ICBS
        & 0.85$\pm$0.09 & 205.50$\pm$17.3 & 0.36$\pm$0.05
        & 0.69$\pm$0.12 & 222.40$\pm$21.7 & 0.41$\pm$0.07
        & 0.52$\pm$0.14 & 248.30$\pm$28.1 & 0.45$\pm$0.08
        & 0.31$\pm$0.16 & 278.60$\pm$34.5 & 0.51$\pm$0.10 \\

        \ \  VORL-A*
        & 0.96$\pm$0.04 & 188.50$\pm$14.8 & 0.33$\pm$0.04
        & 0.88$\pm$0.07 & 202.20$\pm$18.1 & 0.35$\pm$0.05
        & 0.72$\pm$0.11 & 228.40$\pm$24.5 & 0.40$\pm$0.07
        & 0.55$\pm$0.13 & 252.80$\pm$29.2 & 0.44$\pm$0.08 \\

        \ \  VORL-RL
        & 0.86$\pm$0.09 & 203.20$\pm$18.4 & 0.35$\pm$0.05
        & 0.88$\pm$0.08 & 196.40$\pm$16.8 & 0.34$\pm$0.04
        & 0.90$\pm$0.07 & 190.60$\pm$15.5 & 0.33$\pm$0.04
        & 0.92$\pm$0.06 & 193.20$\pm$16.2 & 0.32$\pm$0.04 \\

        \ \  \cellcolor{lightmauve!40}VORL-EXPLORE
        & \cellcolor[HTML]{f3f7fc}\best{0.99}$\pm$0.01 & \cellcolor[HTML]{f3f7fc}\best{181.32}$\pm$12.6 & \cellcolor[HTML]{f3f7fc}\best{0.30}$\pm$0.03
        & \cellcolor[HTML]{f3f7fc}\best{0.98}$\pm$0.01 & \cellcolor[HTML]{f3f7fc}\best{182.71}$\pm$13.1 & \cellcolor[HTML]{f3f7fc}\best{0.31}$\pm$0.03
        & \cellcolor[HTML]{f3f7fc}\best{0.96}$\pm$0.02 & \cellcolor[HTML]{f3f7fc}\best{184.47}$\pm$14.2 & \cellcolor[HTML]{f3f7fc}\best{0.31}$\pm$0.03
        & \cellcolor[HTML]{f3f7fc}\best{0.96}$\pm$0.02 & \cellcolor[HTML]{f3f7fc}\best{188.40}$\pm$14.6 & \cellcolor[HTML]{f3f7fc}\best{0.31}$\pm$0.03 \\

        \bottomrule
    \end{tabular}%
    }
\end{table*}

%% file: table2.tex
\begin{figure}[!htbp]
\centering
\begin{minipage}[t]{0.16\textwidth}
  \centering
  \includegraphics[width=\linewidth]{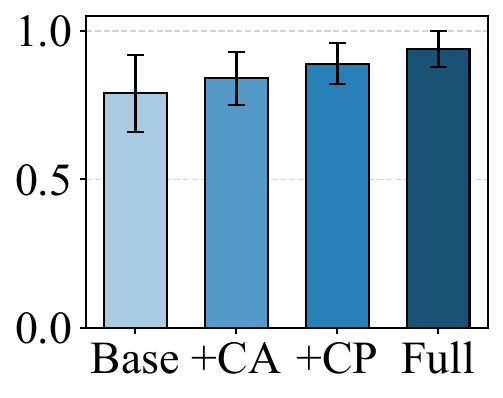}
  \vspace{0.8mm}
  {\small (a) SR}
\end{minipage}\hfill
\begin{minipage}[t]{0.16\textwidth}
  \centering
  \includegraphics[width=\linewidth]{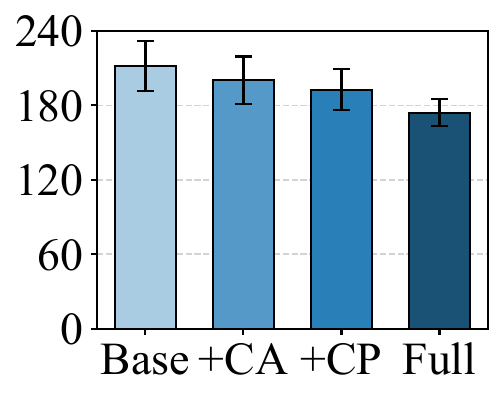}
  \vspace{0.8mm}
  {\small (b) EL}
\end{minipage}\hfill
\begin{minipage}[t]{0.16\textwidth}
  \centering
  \includegraphics[width=\linewidth]{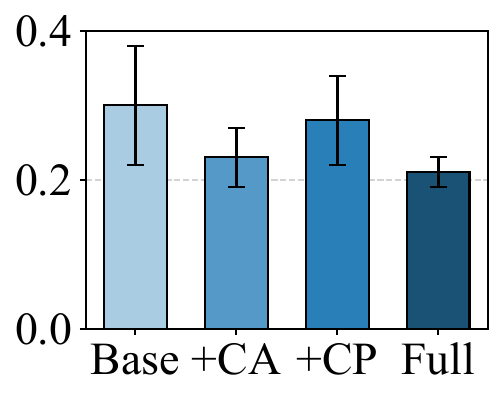}
  \vspace{0.8mm}
  {\small (c) Overlap}
\end{minipage}
\caption{Progressive ablation of the coupled architecture.
Base disables both coupling links and uses decoupled target assignment and motion execution.
CA enables fidelity-coupled assignment by using execution fidelity to reweight frontier scoring, while keeping the execution module unchanged.
CP enables fidelity-gated switching in the execution layer, while keeping the assignment objective unchanged.
Full enables both CA and CP.
Bars show the mean and error bars indicate $\pm$STD over 100 runs with 4 agents on a $40\times40$ grid, 30\% static obstacle density, and 32 dynamic obstacles.}
\label{fig:ablation_results}
\end{figure}